\documentclass[10pt,twocolumn,letterpaper]{article}

\usepackage{cvpr}              

\definecolor{cvprblue}{rgb}{0.21,0.49,0.74}
\usepackage{hyperref}

\def\paperID{7320} 
\def\confName{CVPR}
\def\confYear{2025}

\title{Trajectory Mamba: Efficient Attention-Mamba Forecasting Model Based on Selective SSM}

\author{Yizhou Huang\textsuperscript{1}, 
Yihua Cheng\textsuperscript{2}, Kezhi Wang\textsuperscript{1}\thanks{Corresponding author}\\
\textsuperscript{1}Brunel University of London, \textsuperscript{2}University of Birmingham\\
{\tt\small \{byyh009, Kezhi.Wang\}@brunel.ac.uk, y.cheng.2@bham.ac.uk}\\
\url{https://github.com/YiZhou-H/Trajectory-Mamba-CVPR}
}

\begin{document}
\maketitle
\begin{abstract}
Motion prediction is crucial for autonomous driving, as it enables accurate forecasting of future vehicle trajectories based on historical inputs. This paper introduces Trajectory Mamba, a novel efficient trajectory prediction framework based on the selective state-space model (SSM). Conventional attention-based models face the challenge of computational costs that grow quadratically with the number of targets, hindering their application in highly dynamic environments. 
In response, we leverage the SSM to redesign the self-attention mechanism in the encoder-decoder architecture, thereby achieving linear time complexity.
To address the potential reduction in prediction accuracy resulting from modifications to the attention mechanism, we propose a joint polyline encoding strategy to better capture the associations between static and dynamic contexts, ultimately enhancing prediction accuracy. Additionally, to balance prediction accuracy and inference speed, we adopted the decoder that differs entirely from the encoder. Through cross-state space attention, all target agents share the scene context, allowing the SSM to interact with the shared scene representation during decoding, thus inferring different trajectories over the next prediction steps.
Our model achieves state-of-the-art results in terms of inference speed and parameter efficiency on both the Argoverse 1 and Argoverse 2 datasets. It demonstrates a four-fold reduction in FLOPs compared to existing methods and reduces parameter count by over 40\% while surpassing the performance of the vast majority of previous methods. These findings validate the effectiveness of Trajectory Mamba in trajectory prediction tasks.
\end{abstract}    
\section{Introduction}
\label{sec:intro}
In the rapid development of autonomous driving technology \cite{Liu_2024_CVPR, cheng2024ivgaze, huang2024efficient}, accurate motion prediction plays a crucial role by forecasting future trajectories based on historical vehicle movement data. This task is not only vital for enhancing driving safety but also meets the stringent requirements for real-time system response. In recent years, learning-based approaches have been widely applied to motion prediction \cite{varadarajan2022multipath++, zhou2022hivt, zhou2023query, liang2020learning, gao2020vectornet}. However, existing prediction methods often struggle to balance efficiency and accuracy, making it challenging to simultaneously meet performance and efficiency requirements in real-time applications.
\begin{figure}
    \centering
    \includegraphics[width=\linewidth]{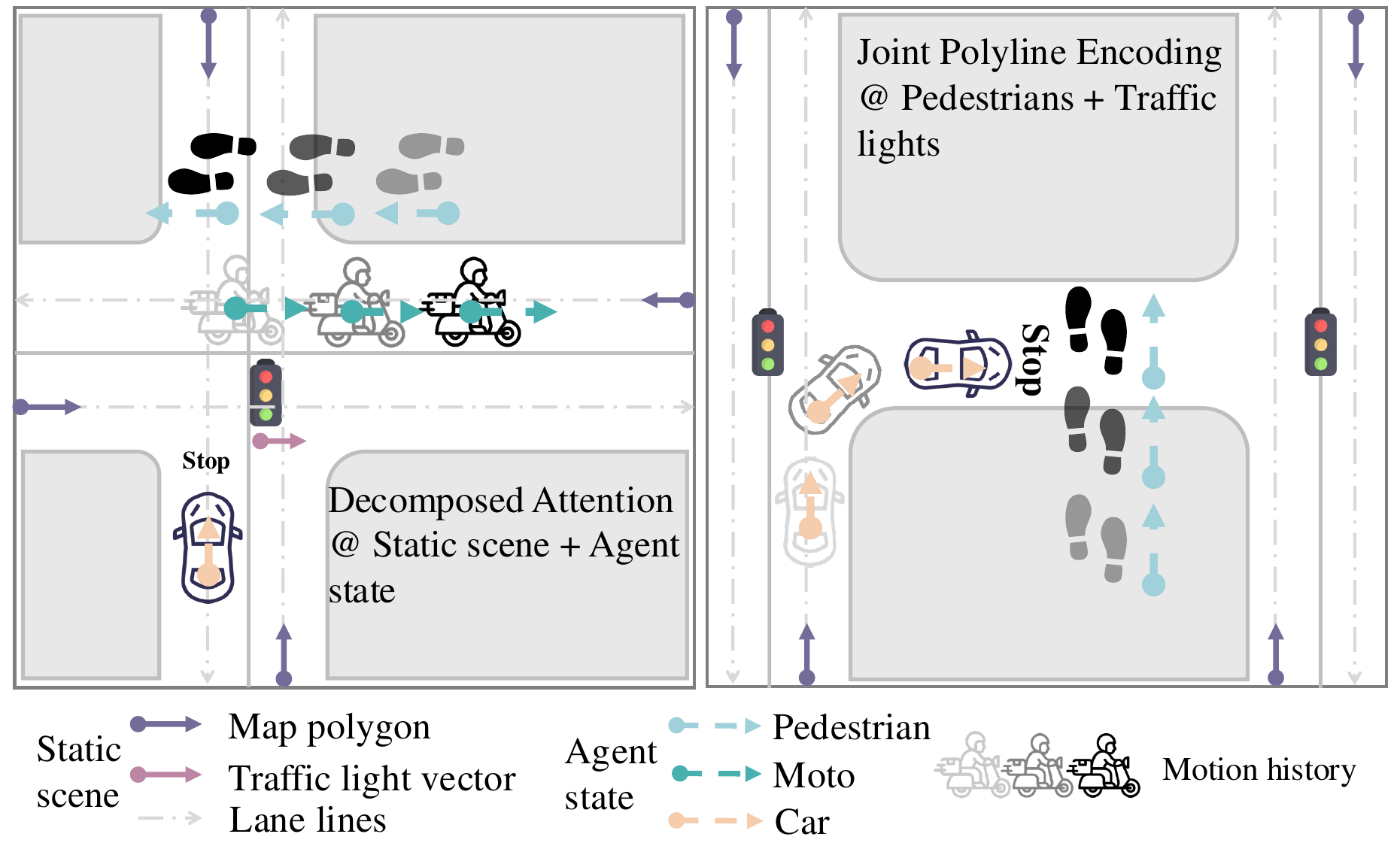}
    \caption{Illustration of proposed joint polyline encoding strategy, where we consider all factors that affect the movement of motor vehicles, jointly encoding pedestrians and traffic lights rather than categorizing the scene horizontally into static and dynamic context. Additionally, we decompose and encode the interactions between all agents and elements at each time step.}
    \label{fig:1}
\end{figure}
We attempt to analyze the underlying reasons from the following aspects and explore possible improvements: 

\textbf{(a). Improving Prediction Accuracy:} In autonomous driving scenarios \cite{zhou2024drivinggaussian}, there are numerous dynamic and static agents, and the diversity of this data presents significant challenges for motion prediction and model training. However, conventional methods \cite{varadarajan2022multipath++, zhou2022hivt} often struggle to effectively fuse this heterogeneous data, particularly when it comes to integrating various features, which leads to suboptimal prediction accuracy.
On a positive note, factorized attention-based transformers \cite{nayakanti2023wayformer, zhou2023query} have made notable strides in decomposing heterogeneous data. However, these models often use a unified, horizontal approach when dealing with static scenes and dynamic agent features, and they fail to fully consider the deeper relationships between different scene elements and agent information.

\textbf{(b). Enhancing Response Efficiency:} Motion predictions are inherently recursive, implying that each prediction serves as an input for the next time step, leading to the progressive accumulation of errors over time. 
Notably, the advancement of sparse context encoding techniques \cite{gao2020vectornet, ye2021tpcn} has demonstrated significant potential in enhancing model efficiency. Nonetheless, prediction models based on attention mechanisms still face exponential growth in computational complexity and parameter size during recursive tasks due to inherent design constraints, which severely limit their efficiency and real-time capabilities in practical applications. Conversely, methods such as LSTM \cite{liu2021social} are relatively more efficient, but numerous studies \cite{mohamed2020social, yuan2021agentformer, salzmann2020trajectron++} have shown that they are incapable of achieving the predictive accuracy of transformer-based models.

In response, we propose an Attention-Mamba framework for motion prediction, termed Trajectory Mamba (Tamba), to overcome the limitations discussed above. First, we design a multimodal encoder-decoder architecture based on the state space model \cite{gu2022parameterization, gupta2022diagonal,smith2024convolutional}, comprising three parallel encoders and a decoder. By introducing a selective state space, we redefined the computation of the self-attention mechanism, enabling the model to selectively process input information by parameterizing the input of the state space model (SSM), thereby allowing the model to focus on or ignore specific inputs. Specifically, the selective state space module processes local short-range features by receiving input from a one-dimensional convolutional layer, ensuring that these features can be selected or discarded before entering the SSM. We utilized local state space attention in place of the multi-head attention in the self-attention module, reducing the computational complexity of recursive reasoning from exponential to linear.

Secondly, to better integrate heterogeneous data features, we proposed a joint polyline encoding strategy. Its objective is to separate pedestrian agents from the dynamic agents and jointly encode them with traffic light information from the static scene (see Fig. \ref{fig:1}). We observed that in uncertain motion prediction scenarios, the application of traffic regulations exerts a degree of implicit control over the potential movement states of agents. For example, pedestrian-priority traffic rules remain effective even in the absence of traffic lights, influencing vehicles and motorcycles alike. By jointly encoding pedestrians and traffic lights through a shared embedder, this jointly encoding strategy can independently interact with other dynamic agents to enhance the prediction accuracy of vehicle and motorcycle agents.

Lastly, we designed a prediction weight inference module. By redesigning selective state space with a cross-attention mechanism, we utilized independent query tensors to recursively infer multimodal candidate trajectories through interactions with the encoder and the keys and values from the previous recursive step. These candidate trajectories were then evaluated using a recurrent neural network to assign trajectory scores.

Our contributions are summarized as: (1) We propose the Tamba framework for motion prediction, which leverages the state space model to redesign computations of self-attention, resulting in computational complexity reducing significantly. 
(2) We propose a joint polyline encoding strategy. By integrating strongly correlated polyline types using independent or shared embedders, elements that impose similar behavioral patterns can better influence different agents, collectively contributing to the trajectory reasoning of these agents.
(3) We redesign the Mamba decoder with cross-attention, allowing the joint scene encoding to share a unified scene representation, which further reduces the model's complexity. At the final stage, we introduced a prediction weight inference module to refine the predicted trajectories and enhance prediction accuracy.
(4) Experimental results demonstrate that our method achieves state-of-the-art (SOTA) performance on Argoverse 1 and 2 datasets. Compared to advanced methods, our method reduces the number of parameters by 40\%, achieves a four-fold increase in FLOPs. This represents a new level of balance between efficiency and performance in our model. 

\section{Related Works}
\label{sec:formatting}

\subsection{State Space Model}
Recently, the Mamba \cite{gu2023mamba} framework has rekindled interest in state-space models (SSMs) \cite{gu2022parameterization, gupta2022diagonal,smith2024convolutional} due to its introduction of an input-dependent selection mechanism, which enhances their applicability and performance. SSMs are renowned for their capacity to handle long-range dependencies and offer superior memory and computational efficiency compared to transformer-based architectures \cite{vaswani2017attention}. Hippo \cite{gu2020hippo}, a seminal contribution to SSMs, established the efficacy of these models in long-range sequence modeling tasks. Subsequent research, such as Vision mamba \cite{zhu2024vision}, has further refined the mechanisms underlying SSMs, resulting in substantial performance improvements. SSMs have demonstrated their versatility and robustness across a diverse set of domains, including computer vision, video analysis, medical imaging, graph-based learning, point cloud processing, and recommendation systems.

In the domain of trajectory prediction, recent studies have indicated that Motion Mamba \cite{zhang2025motion} achieves notable success in predicting human trajectory time series. Building on these advancements, this work integrates the state-space mechanisms of Mamba with attention mechanisms to address trajectory prediction tasks in dynamic traffic scenarios, thereby leveraging both the efficiency of SSMs and the adaptability of attention-based models.

\subsection{Attention Trajectory Prediction Model}
In the field of trajectory prediction, methods \cite{kothari2021interpretable, mangalam2021goals} based on the attention mechanism have been widely adopted to enhance the model's ability to capture and predict complex behaviors. These approaches leverage the capability of attention to focus selectively on important parts of the input, thereby capturing key temporal and spatial features that are crucial for predicting future trajectories.

Subsequently, several advanced approaches \cite{Wong_2024_CVPR, Liu_2024_CVPR} have utilized attention mechanisms for trajectory prediction specifically in multi-agent interaction environments, where understanding the interactions between different agents (such as vehicles, pedestrians, and other road users) is critical. These models employ social attention modules to focus on the movements and behaviors of surrounding vehicles, effectively improving the ability of the system to handle intricate interactive behaviors, which are a common aspect of real-world driving scenarios.

Moreover, attention mechanisms have also been successfully applied in models that integrate multimodal information \cite{shi2023representing, shi2024mtr++, zhou2023query}. These models dynamically weight and fuse different types of sensory inputs—such as image features from cameras and data from other sensors like LiDAR—through attention modules. This fusion process allows the model to better understand and represent complex traffic environments, effectively utilizing complementary data to enhance trajectory prediction performance.

\section{Our Method}
\subsection{Overview}
In the autonomous driving scenario, the input consists of the predicted dynamic agent, its surrounding agents, and the static scene context. Let the historical state of the $i$-th agent at time step $t$ be represented as: $s_i^t = (p_i^t, \theta_i^t, t, v_i^t)$, where $p_i^t = (p_{i,x}^t, p_{i,y}^t)$  denotes the spatial position,  $\theta_i^t$ represents the heading angle, $t$ refers to the time step, and $v_i^t$  denotes the velocity.
The static scene context is represented by $n$ polygons, which include elements such as lane boundaries, map edges, and sidewalks.
The agent states $s_i^t$ over $T$ time steps within the observation window are used to predict $K$ future trajectories over $T^{\prime} $ time steps.

In this paper, we propose the Tamba model to optimize the efficiency and accuracy of motion forecasting tasks. Our input includes $s_i^t$ of dynamic and static agents, $\mathcal{P}$ processed using a heterogeneous polyline encoding strategy to transform $s_i^t$ into higher-dimensional features. We leverage three Tamba encoders to seperately decompose ($\mathrm{I}$) attentations in each time step ($t$) over all elements, ($\mathrm{II}$) attentations between all agents (A) and scene (S), and ($\mathrm{III}$) attentions of traffic control elements (pedestrians and traffic lights) to other dynamic agents (motorcycles and cars). Then, concate features from encoders as output to a Tamba decoder, where we utilize cross-attention mechanism to generate proposal trajectories. Subsequently, we fused scene encoding with historical information to perform a secondary decoding of the proposal trajectories for refinement. Finally, the refined trajectories were optimized using a “winner-takes-all strategy” to obtain trajectory output.

Our model contains a novel heterogeneous polyline encoding strategy along with a decomposed feature fusion method for both static and dynamic agents. A redesigned Tamba encoder-decoder architecture. A refinement process to output the final predicted trajectory. In the following sections, we sequentially introduce the components in Tamba.

\begin{figure}
    \centering
    \includegraphics[width=\linewidth]{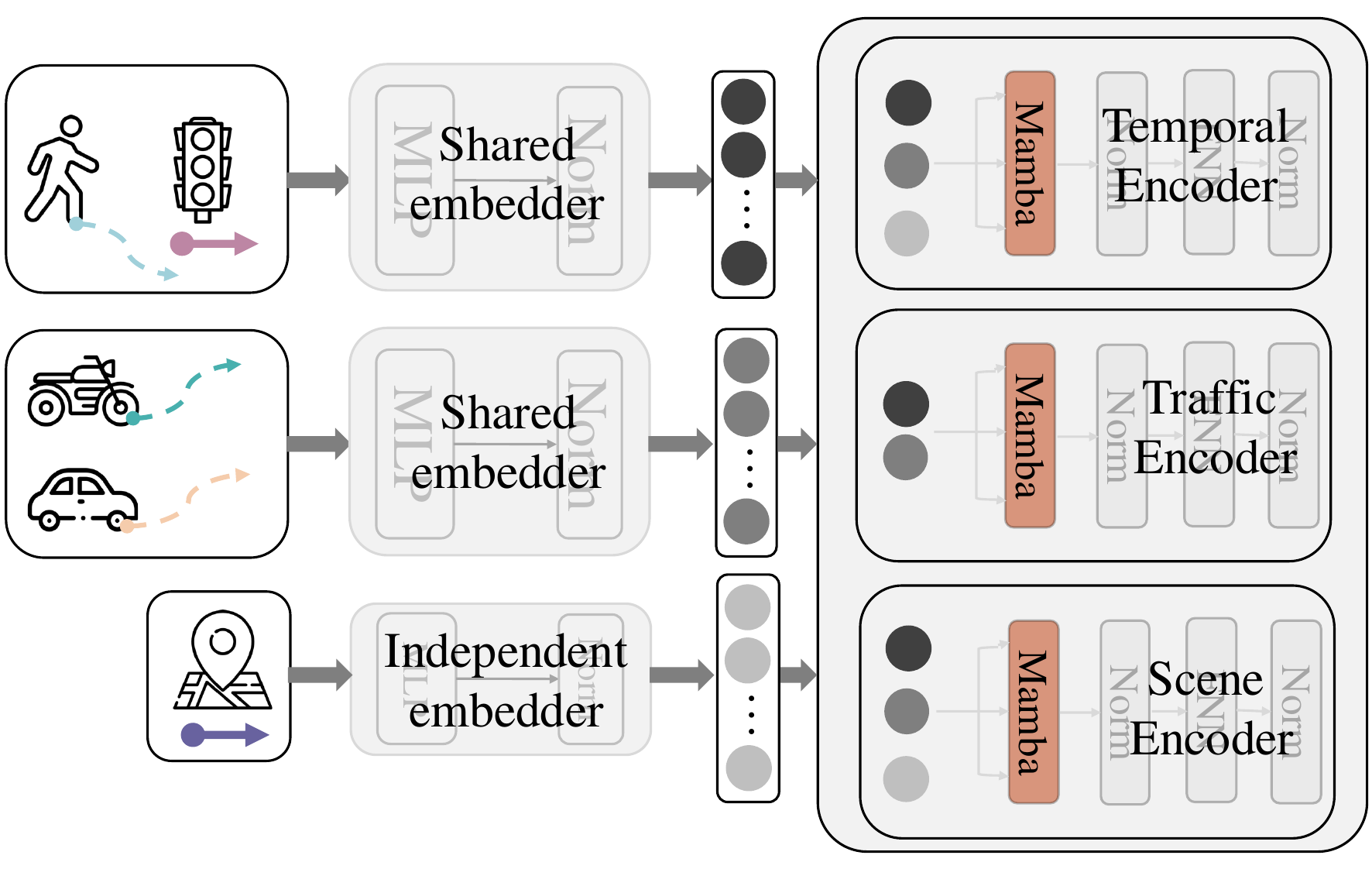}
    \caption{Overview of our Tamba encoder. We employ joint polyline encoding strategy to integrate strongly correlated polyline information and used three parallel encoders to interact with and associate these features. We applied a linear projection to map the attention output of Tamba back to the same dimensions as the input and used normalization to enhance the stability of the output. }
    \label{fig:2}
\end{figure}

\subsection{Joint Polyline Encoding}
In multimodal traffic environment prediction tasks, different types of polylines possess unique geometric characteristics and semantic information. This study proposes a novel heterogeneous polyline embedding method to facilitate the fusion of dynamic and static context features, thereby providing a richer and more efficient feature representation. Specifically, we assign either independent or shared embedders to each type of polyline to effectively model various categories of information. Specifically, we jointly encode polylines with strong information associations, such as pedestrian trajectories and traffic light signals, using a shared embedder. This is because agents (pedestrians) and traffic lights typically impose similar behavioral patterns on other agents (vehicles) and often participate together in traffic control. Thus, using a shared embedder reduces model complexity and promotes feature sharing, collectively influencing vehicle trajectory reasoning.

In contrast, independent embedders are used for polyline types with significant feature differences, such as lane lines and traffic signs in maps. Each embedder consists of a MLP layer and normalization layers to extract high-dimensional feature representations. The shared pedestrian and traffic light embedder uses a cross-category feature fusion layer to enhance the model's sensitivity to different information sources. Polylines retain sequential information through positional encoding before being input to the embedders for feature extraction, followed by cross-category feature interaction via the Tamba encoder.

The mathematical representation of this process is as follows: Given the individual polyline data $P = \{p_1, p_2, \dots, p_n\}$, where $p_i \in \mathbb{R}^d$ represents a point on the polyline, $n$ is the number of points, and $d$ is the dimensionality, the embedding process is represented as: $\mathcal{P} = \text{Embed} (P)$. For the pedestrian and traffic light categories that share an embedder, the output feature representation $\mathcal{P}_{\text{joint}}$ is computed as follows:
\begin{equation}
\label{deqn_ex1}
\mathcal{P}_{\text{joint}} = \text{Fusion}\left(\text{Embed}\left(\mathcal{P}_{\text{pedestrian}}\right), \text{Embed}\left(\mathcal{P}_{\text{traffic}}\right)\right),
\end{equation}
where $\mathcal{P}_{\text{pedestrian}}$ and $\mathcal{P}_{\text{traffic}}$ are the input polylines for pedestrians and traffic lights, respectively, and \text{Fusion} represents the feature fusion operation.

\subsection{Attention Tamba Encoder}
In the proposed encoder, the input features $\mathcal{P}$ are uniformly embedded through joint polyline encoding. The input feature shape of the Tamba encoder is ($n, L, d$), where $n$ denotes the number of encoded agents or scene elements, $L$ represents the temporal sequence length, and $d$ indicates the feature dimensionality at each time step.

Specifically, the sequence ($L, d$) undergoes linear transformations to generate the query, key, and value matrices, each with dimensions ($L, d_k$), where $d_k$ is the feature dimensionality used in the attention mechanism. The dot-product similarity between the query and key matrices is computed to yield an attention weight matrix of shape ($L, L$), which captures the interdependencies between all time steps. Subsequently, these attention weights are employed to compute a weighted sum of the value matrix, yielding an output of shape ($L, d_k$). 

As the output feature dimensionality of the Tamba block ($d_k$) may differ from the input feature dimensionality ($d$), a linear projection is employed to align the attention mechanism’s output with the input dimensionality, resulting in ($L, d$). This alignment ensures compatibility for the residual connection. Subsequently, the original input and the linearly projected attention output are combined via element-wise addition, yielding a resultant tensor of shape ($L, d$). To further stabilize the output, a normalization operation is applied, preserving the same dimensionality of ($L, d$).

Subsequently, the features are processed by a feedforward neural network comprising two fully connected layers. The first layer projects the input to a higher-dimensional space $d_{ff}$, resulting in an output with dimensions ($L, d_{ff}$). The second layer then maps the expanded features back to the original dimensionality $d$, restoring the shape to ($L, d$). The output of the feedforward network is combined with the original input through a residual connection, preserving the output shape of ($L, d$). A normalization operation is then applied to ensure stability, maintaining the final output shape of ($L, d$) for the entire encoder module. Thus, within the encoder module, the temporal sequence length ($L$) remains constant, whereas the feature dimensionality ($d$) may briefly change during processing but ultimately returns to its initial dimensionality.

Unlike traditional transformer models, Tamba adjusts the parameters of its state space model (SSM) based on the input data, allowing it to selectively retain or disregard different pieces of information within the sequence. Mathematically, Tamba uses a state space model to capture the temporal evolution of the input. Given an input sequence $\mathcal{P} = \{\mathcal{P}_1, \mathcal{P}_2, \dots, \mathcal{P}_T\}$ , where $T$ represents the sequence length and $\mathcal{P}_t \in \mathbb{R}^d$ is the input at each time step $t$ , the SSM is governed by the following state evolution equation:
\begin{equation}
\label{deqn_ex2}
\mathbf{h}_{t+1} = A(\mathcal{P}_t) \mathbf{h}_t + B(\mathcal{P}_t) \mathbf{u}_t,
\end{equation}
Here $\mathbf{h}_t \in \mathbb{R}^n$ represents the hidden state at the time step $t$, while $A(\mathcal{P}_t)\in \mathbb{R}^{n \times n}$ and $B(\mathcal{P}_t) \in \mathbb{R}^{n \times m}$ are matrices that depend on the input and evolve over time, and $\mathbf{u}_t \in \mathbb{R}^m$ represents the control input.

The output of the time step $t$ is denoted as $\mathbf{y}_t \in \mathbb{R}^p$, and is computed as follows:
\begin{equation}
\label{deqn_ex3}
\mathbf{y}_t = C(\mathcal{P}_t) \mathbf{h}_t + D(\mathcal{P}_t) \mathbf{u}_t.
\end{equation}
where $C(\mathcal{P}_t)\in\mathbb{R}^{p\times n}$ and $D(\mathcal{P}_t) \in \mathbb{R}^{p \times m}$ are also matrices that depend on the input. This formulation allows the system to dynamically adjust its parameters based on the current input, facilitating the selective retention and forgetting of information. The computational complexity of this method is linear, as the state update and output computation at each time step involve matrix-vector multiplication. The overall complexity is $O(T \cdot n^2)$, where $T$ represents the sequence length and $n$ is the state dimension.

\begin{figure*}
    \centering
    \includegraphics[width=\linewidth]{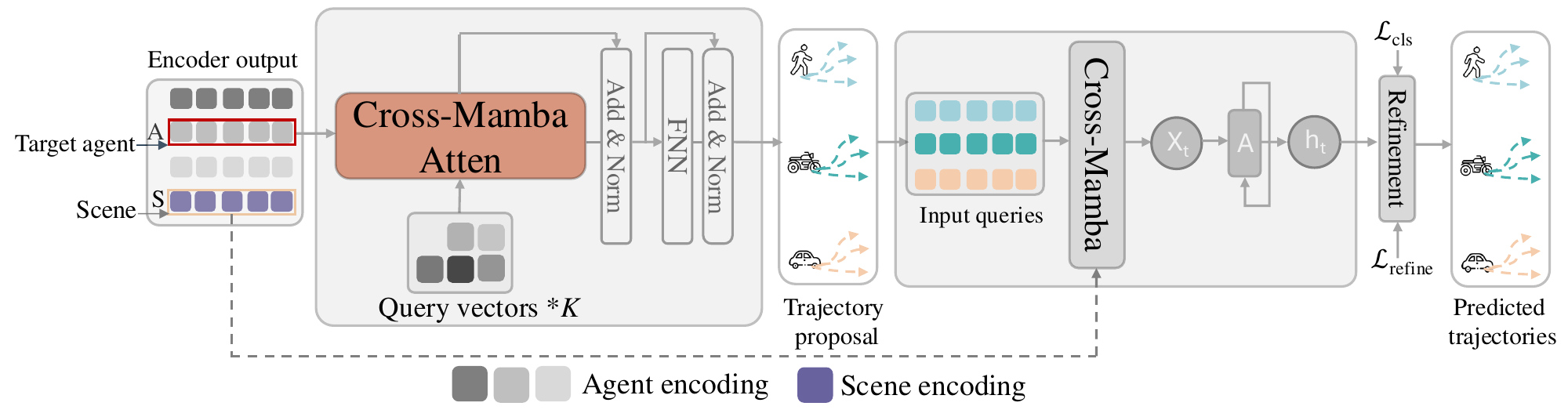}
    \caption{Overview of our Tamba decoding process. The outputs of the three Atten-Tamba encoders carry the context of both the static scene and dynamic agent states, which are concatenated and input into the Cross-Tamba decoder. For predicting the target agent's trajectory, the decoder queries $K$ trajectory modes using independent query tensors, while the encoder's features combine the current state with recursive reasoning from the previous step, allowing all target agents to share a unified scene representation. During the secondary decoding, the predicted state of the proposed trajectory interacts once again with the current scene information, and a recurrent network assigns prediction weights to the proposed trajectories. The final output trajectory is obtained through refinement of these predictions.}
    \label{fig:3}
\end{figure*}

\textbf{Decomposed Attention Mechanism.} We redesigned the self-attention computation, replacing the multi-head attention mechanism with state space model. After receiving the polyline encoding output $\mathcal{P}$, Tamba performs the attention decomposition using three parallel encoders. The detailed computation process is as follows:

($\mathrm{I}$) Spatiotemporal attention. The spatiotemporal attention for each agent $A_i$ and scene elements (S) over the $\mathcal{P}$ at each time step $t$ can be represented as: $Z_\text{time} = \text{softmax}\left( \frac{Q_t K_t^\top}{\sqrt{d_k}} \right) V_t$, where $Q_t, K_t, V_t\in\mathbb{R}^{T\times d_k}$ are the query, key, and value matrices at time step $t$ respectively, and $d_k$ represents the dimensionality of the key vectors. 

($\mathrm{II}$) Scene attention. For the attention between agent set $A$ and scene elements $S$, let $A_i\in \mathbb{R}^{d_a}$ represent the embedding of agent $i$, and $S_j\in \mathbb{R}^{d_m}$ represents the embedding of scene element $j$. The attention calculation between agent $i$ and scene element $j$ is given by: $\alpha_{ij} = \frac{\exp(A_i^\top S_j)}{\sum_{k} \exp(A_i^\top S_k)}$, where $\alpha_{ij}$ denotes the attention weight, indicating the relevance of scene elements $j$ to agent $i$. The output is a weighted sum of the scene elements features denote as:
\begin{equation}
\label{deqn_ex4}
Z_\text{scene} = \sum_j \alpha_{ij} \text{S}_j,
\end{equation}

($\mathrm{III}$) Traffic Attention. Similar to Eq. \ref{deqn_ex4}, we use $\beta_{ij}$ to represent the attention weight of traffic control elements (pedestrians and traffic lights) to other dynamic agents (e.g., motorcycles and cars). For a agent $i$, the output feature of traffic attention can be expressed as: $Z_\text{traffic} = \sum_j \beta_{ij} A_j$. This output is obtained by calculating the weighted sum of the embedded vectors $A_j$ of all dynamic agents, where the weight $ \beta_{ij}$ reflects the impact of the traffic control elements on the dynamic agents. Through this interaction, the model can effectively focus on the influence of pedestrians on dynamic agent movement in non-traffic light scenarios.

\subsection{Cross Tamba Decoder}
The core challenge of trajectory decoding lies in addressing the ‘one-to-many’ problem, that is, how to generate $K$ output results (possibilities) from one set of input features. In designing the decoder architecture, we drew inspiration from DETR’s approach \cite{carion2020end} to solving the ‘one-to-many’ problem in object detection tasks. Specifically, the number of $K$ query vectors is initialized to represent different objects or candidate future trajectories. Through multiple iterations of the attention mechanism, each query vector extracts relevant information from the input features to produce $K$ output results.

\textbf{Crossed Dynamic and Static Attention.} In the proposed cross-Tamba decoder, we designed it by using state space model to replace multi-head attention in the cross-attention mechanism. The query vector $Q$ is separated to interact independently with the keys and values from the encoder’s output features. Specifically, the query vector is responsible for querying the modes of $K$ trajectories, while $K$ and $V$ incorporate both the encoded features from the encoder and the reasoning from the previous recursive step. This design allows dynamic agents (e.g., vehicles and motorcycles) to better share mixed features (pedestrians and traffic lights), while all target agents share a single static scene representation and iteratively generate $K$ different proposed trajectories over $T^{\prime}$ future recursive steps. During the training phase, the proposed trajectories are optimized using the MSE loss ($L_{\text{proposal}}$) based on the provided labels.

\textbf{Trajectory Proposal.} After proposing $K$ predicted trajectories, we introduce a prediction weight inference module to compute the weight for each trajectory. In this process, we also use the Cross-Tamba module, which first fuses scene (S) with historical information to infer the likelihood of each predicted trajectory. Specifically, given each trajectory  $p_k$  (where  $k = 1, 2, \ldots, K $), we can represent it as:
\begin{equation}
\label{deqn_ex5}
p_k = \text{Cross-Tamba}(\text{Scene}, \text{Predictions}).
\end{equation}
where $\text{Scene}$ represents the feature representation of the current scene and $\text{Predictions}$ contains all predicted proposal trajectories.
At the head of the weight inference module, we use a recurrent neural network (RNN) \cite{sherstinsky2020fundamentals} to extract features from the final trajectories in order to generate a score for each trajectory. This RNN module can be represented as:
\begin{equation}
\label{deqn_ex6}
h_t = \text{RNN}(h_{t-1}, p_k),
\end{equation}
where $h_t$ is the hidden state at time step $t$, $h_{t-1}$ is the previous state. Through this process, the RNN provides a mechanism for constraints across trajectories. When a trajectory conflicts with the environment or when there is an alternative trajectory that better fits the overall distribution, the RNN generates a prior confidence score for these trajectories. The confidence generated by the RNN can be represented as: $C_k = f(h_T)$, where $C_k$ is the confidence score for trajectory $p_k$, $h_T$ is the hidden state of the RNN at the final time step, and $f$ is a nonlinear activation function.

This step ensures greater stability in the optimization process, such that the proposal trajectory considers not only the current prediction results but also the contextual environment and historical information, thereby enhancing the overall performance and robustness of the model.

\textbf{Trajectory Refinement.}
We employed the same optimization approach as the baseline methods \cite{zhou2022hivt, zhou2023query} to perform mixture modeling of the predicted trajectories. The future trajectory of agent $i$ is modeled as a mixture of Laplace distributions, as given by:
\begin{equation}
\label{deqn_ex7}
f(\{p_{t}^{i}\}_{t=1}^{T'}) = \sum_{k=1}^{K} \pi_{i,k} \prod_{t=1}^{T'} \text{Laplace}(p_{t}^{i} | \mu_{t,i,k}, b_{t,i,k}),
\end{equation}
where $\{ \pi_{i,k} \}_{k=1}^{K}$ are the mixing coefficients, representing the significance of each mixture component in the overall model. For each time step $t$, the Laplace density of the $k^{th}$ mixture component is defined by the location parameter $\mu{t,i,k}$ and the scale parameter $b_{t,i,k}$, which reflect the center and spread of that component.
We use a classification loss $L_{\text{cls}}$ to optimize the mixing coefficients in Eq. \ref{deqn_ex7} by minimizing their negative log-likelihood. It is important to note that during this process, we stop the gradients of the location and scale parameters, focusing solely on the optimization of the mixing coefficients. This allows us to effectively focus on the classification task without affecting the trajectory’s location and scale settings.

For optimizing the location and scale parameters, we adopt the “winner takes all” strategy \cite{lee2016stochastic}, resulting in only the best-predicted trajectory from the proposal and refinement modules undergoing backpropagation. This approach ensures that the network is adjusted only based on the best predictions during training, thereby enhancing the accuracy and stability of the model. In addition, to prevent overfitting and improve model robustness, the refinement module halts the gradient propagation of the proposed trajectory. This strategy helps to effectively preserve the previously learned information during the adjustment of proposals.

Finally, our loss function integrates three components: the trajectory proposal loss ($L_{\text{proposal}}$), the trajectory refinement loss ($L_{\text{refine}}$), and the classification loss ($L_{\text{cls}}$), enabling an end-to-end training process. The expression is given as:
\begin{equation}
\label{deqn_ex8}
L_{\text{total}} = L_{\text{proposal}} + L_{\text{refine}} + \lambda L_{\text{cls}}.
\end{equation}

\begin{table*}[t]
\scriptsize
\centering
\fontsize{9pt}{11pt}\selectfont
\tabcolsep=0.15cm
\renewcommand\arraystretch{1.1}
\begin{tabular}{c|cccc|ccc|cc}
\hline
Method & \textbf{b-minFDE$_{6}$}$\downarrow$ &minADE$_{6}$$\downarrow$ & minFDE$_{6}$$\downarrow$ & MR$_{6}$$\downarrow$ & minADE$_{1}$$\downarrow$ & minFDE$_{1}$$\downarrow$ & MR$_{1}$$\downarrow$ & \#Params (M)$\downarrow$&FLOPs(G)$\downarrow$ \\
\hline
THOMAS \cite{gilles2022thomas} & 2.16 & 0.88 & 1.51 & 0.20 & 1.95 & 4.71 & 0.64 & - & -\\
GoRela \cite{cui2023gorela} & 2.01 & 0.76 & 1.48 & 0.22 & 1.82 & 4.62 & 0.66 & - & -\\
QML \cite{su2022qml} & 1.95 & 0.69 & 1.39 & 0.19 & 1.84 & 4.98 & 0.62 & 9.39 & -\\
MTR \cite{shi2024mtr++} & 1.98 & 0.73 & 1.44 & 0.15 & 1.74 & 4.39 & 0.58 & 65.78  & - \\
GANet \cite{wang2023ganet} & 1.96 & 0.72 & 1.34 & 0.17 & 1.77 & 4.48 & 0.59 & 61.73  & 15.79 \\
BANet \cite{wang2023technical} & 1.92 & 0.71 & 1.36 & 0.19 & 1.79 & 4.61 & 0.60 & 9.49  & \textbf{11.93} \\
QCNet \cite{zhou2023query} & 1.91 & 0.65 & 1.29 & \textbf{0.16} & 1.69 & 4.30 & 0.59 & 7.66 & 45.3\\
\hline
\textbf{Tamba (Ours)} & \textbf{1.89} & \textbf{0.64} & \textbf{1.24} &0.17 & \textbf{1.66} & \textbf{4.24} & \textbf{0.57} & \textbf{4.54} & 27.3\\
\hline
\end{tabular}
\caption{Performence comparison on Argoverse 2 \cite{Argoverse2, TrustButVerify} dataset, leaderboard ranked by b-minFDE$_{6}$. Subscript $_{6}$ and $_{1}$ represents the number of predictions. Best performance is highlighted in \textbf{bold}. (M) represents the number of parameters in millions; (G) refers to GigaOps per second. - represents we were unable to reproduce the results with complete accuracy. Noteably, all methods in the table are direct comparisons of the original approaches. We exclude additional improvements from engineering techniques such as ensembling.}
\label{tab:ag1}
\end{table*}

\begin{table*}[t]
\fontsize{9pt}{11pt}\selectfont
\centering
\renewcommand\arraystretch{1.1}
\begin{tabular}{c|cccc|c}
\hline
Method & \textbf{b-minFDE$_{6}$}$\downarrow$ & minADE$_{6}$$\downarrow$ & minFDE$_{6}$$\downarrow$ & MR$_{6}$$\downarrow$ & \#Params (M)$\downarrow$ \\
\hline
LaneGCN \cite{liang2020learning} & 2.06 & 0.87 & 1.36 & 0.16 & - \\
DenseTNT \cite{gu2021densetnt} & 1.98 & 0.88 & 1.28 & 0.13 & - \\
MTR \cite{shi2024mtr++} & 1.93 & 0.82 & 1.24 & 0.13 & - \\
SceneTransformer \cite{ngiamscene} & 1.89 & 0.80 & 1.23 & 0.13 & 15.30 \\
HiVT \cite{zhou2022hivt} & 1.84 & 0.77 & 1.17 & 0.13 & - \\
GANet \cite{wang2023ganet} & 1.79 & 0.81 & 1.16 & 0.12 & - \\
Wayformer \cite{nayakanti2023wayformer} & 1.74 & 0.77 & 1.16 & 0.12 & - \\
QCNet \cite{zhou2023query}& 1.69 & 0.73 & \textbf{1.07} & 0.11 & 7.66 \\
\hline
\textbf{Tamba (Ours)} & \textbf{1.67} & \textbf{0.72} & \textbf{1.07} & \textbf{0.09} & \textbf{4.54} \\
\hline
\end{tabular}
\caption{Quantitative results on the Argoverse 1 \cite{Argoverse} dataset, leaderboard ranked by b-minFDE$_{6}$. Subscript $_{6}$ represents the number of predictions. Best performance is highlighted in \textbf{bold}. (M) represents the number of parameters in millions. - represents it is unable to reproduce the results with complete accuracy. We exclude additional improvements from engineering techniques such as ensembling.}
\label{tab:ag2}
\end{table*}

\section{Experiments}
\subsection{Datasets}
Our experiments were conducted on the Argoverse 1 \cite{Argoverse} and Argoverse 2 \cite{Argoverse2, TrustButVerify} datasets. It contains extensive real-world autonomous driving scenarios and are designed to advance the development of autonomous driving technologies. The Argoverse 1 dataset includes 360-degree sensor data from the Pittsburgh and Miami regions in the United States, covering various road types and complex urban environments. Sampled at a rate of 10Hz, this dataset provides trajectory data of the ego vehicle over a 5-second time span, with 2 seconds of observation used to predict the path over the next 3 seconds, comprising over 30,000 such samples.
Argoverse 2 further expands on the dataset by including richer sensor data, multimodal annotations, and extending the observation window. Specifically, it includes 11-second scenarios, with 5 seconds of observation followed by prediction of trajectories for the next 6 seconds. The dataset contains 250,000 scenarios and features a large object taxonomy comprising 10 non-overlapping classes, which is used for training and validating motion prediction models.

\subsection{Metrics}
We adopted evaluation methods consistent with prior research, using metrics such as minADE($_K$), minFDE($_K$), b-minFDE($_K$), and MR($_K$) to measure model performance in terms of error. Specifically, minADE($_K$) calculates the average $L2$ distance between the ground truth and the most optimal trajectory among the $K$ predicted trajectories across all future time steps, thereby assessing the overall prediction accuracy of the model over multiple time steps. minFDE($_K$) focuses on the prediction error at the final time step, measuring the model’s ability to predict the final state accurately.
b-minFDE($_K$) extends minFDE($_K$) by incorporating probability weighting, using $(1 - \hat{\pi})^2$ as a weight to account for prediction uncertainty, where $\hat{\pi}$ represents the probability score assigned by the model to the optimal predicted trajectory. MR($_K$) measures the proportion of the $K$ predicted trajectories with an error greater than 2 meters, providing an assessment of the model’s stability and the frequency of prediction failures. Together, these metrics offer a comprehensive evaluation of the model’s prediction accuracy, stability, and uncertainty estimation.
For the parameter $K$, we followed conventional values of 6 and 1. When the number of predicted trajectories exceeds $K$, the top $K$ trajectories with the highest probabilities are selected. In terms of lightweight evaluation, we used the number of parameters (M) and FLOPs(G) as the metrics.

\begin{figure*}
    \centering
    \includegraphics[width=\linewidth]{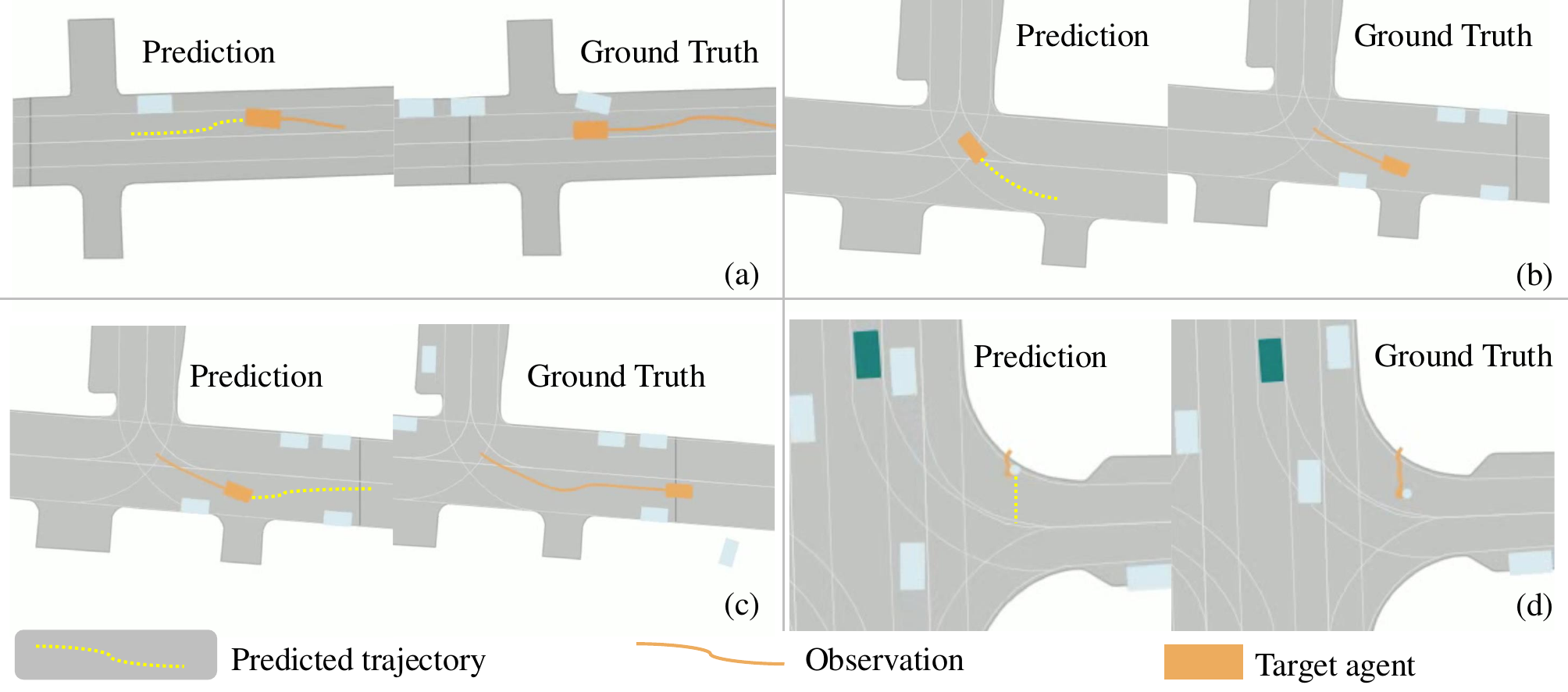}
    \caption{Qualitative results on the Argoverse 2 validation set. The target agent marked as orange with its surrounding agents marked as cold white. We conduct four different traffic scenarios. (a). Multi-agents in straight-road scenario, (b). Multiple agents on roundabout road, (c). Vehicle avoidance after roundabout, (d). Mixed scenario involving pedestrians and vehicles.}
    \label{fig:4}
\end{figure*}
\subsection{Settings}
We conducted all experiments under the same environment (batch size, epochs) using eight NVIDIA 3090TI cards with a batch size of 128. We used Adam for optimization, with an initial learning rate of 0.001. Training was conducted for a total of 50 epochs. We employed an adaptive learning rate adjustment strategy based on the validation set's performance during training. Specifically, when the validation accuracy did not improve for five consecutive epochs, the learning rate was reduced to 0.1 times its original value.

\begin{table*}[htbp]
\fontsize{9pt}{11pt}\selectfont
\renewcommand\arraystretch{1.1}
\setlength{\tabcolsep}{2pt} 
\centering
\begin{tabular}{c|cc|cc|cc|cc}
\hline
\multirow{2}{*}{Method} & \multicolumn{4}{c|}{w/ joint} & \multicolumn{4}{c}{w/o joint} \\
\cline{2-9}
 & minFDE\textsubscript{6}$\downarrow$ & minADE\textsubscript{6} $\downarrow$& minFDE\textsubscript{1}$\downarrow$ & minADE\textsubscript{1}$\downarrow$ & minFDE\textsubscript{6}$\downarrow$ & minADE\textsubscript{6}$\downarrow$ & minFDE\textsubscript{1}$\downarrow$ & minADE\textsubscript{1}$\downarrow$ \\
\hline
Ours + Attention & 1.094 & 0.763 & 3.037 & 1.376 & 1.093 & 0.728 & 3.035 & 1.372 \\
Ours + Mamba     & 1.091 & 0.724 & 2.899 & 1.298 & 1.083 & 0.765 & 2.821 & 1.292 \\
\textbf{Ours + Tamba}     & \textbf{0.951} & \textbf{0.673} & \textbf{2.742} & \textbf{1.265} & \textbf{0.955} & \textbf{0.680} & \textbf{2.745} & \textbf{1.272} \\
\hline
\end{tabular}
\caption{Ablation study on the components of the Tamba encoder-decoder. Experimental results are conducted on the Argoverse 2 \cite{Argoverse2, TrustButVerify} validation set.}
\label{tab:methods_performance_joint}
\end{table*}

\subsection{Comparison with State of the Art}
We compare our method with other SOTA approaches across various evaluation metrics in Argoverse 2 dataset. As shown in Table \ref{tab:ag1}, our method demonstrates superior performance on key metrics such as minFDE$_{6}$ and minADE$_{6}$. Tamba achieves a b-minFDE$_{6}$ score of 1.79, compared to 1.91 for QCNet, 1.92 for BANet and 1.98 for MTR. Additionally, Tamba achieves (MR$_{1}$) of 0.57, which is lower than that of other methods, such as QCNet (0.59) and MTR (0.58), indicating more stable prediction accuracy. Furthermore, Tamba has a parameter count of 4.54 M, which is significantly lower than QCNet (7.66 M), even compare to non-transformer models such as BANet (9.49M) and GANet (61.73M), Tamba achieving reductions of 52.0\% and 92.7\% respectively, while still outperforming both models. Additionally, Tamba achieves 27.3 GigaOps per second, which is 40\% faster than QCNet’s 45.3 GigaOps per second. These results clearly demonstrate that Tamba is competitive in terms of accuracy but also excels in lightweight design and efficiency, providing an excellent solution for complex task.

On the Argoverse 1 dataset, our method outperforms other state-of-the-art (SOTA) methods across the board. On the leaderboard for the b-minFDE$_6$ metric,Tamba achieves a score of 1.67, while the second-best, QCNet, scores 1.69, and the third, BANet, scores 1.74. In terms of parameter count, our method requires 4.54M parameters, compared to 7.66M for QCNet, reflecting a 40\% reduction. This further demonstrates the superiority of our approach in balancing efficiency and performance. For detailed performance metrics, please refer to Table \ref{tab:ag2}.

\subsection{Quantization Results}
We visualized the prediction results on the validation set. In some challenging sequences, the predictions made by the Tamba model are more reasonable and smoother compared to existing methods, effectively adhering to map constraints. In Figure \ref{fig:4}, we present several multimodal prediction cases, the results demonstrate that the trajectories generated by Tamba are smoother and better adapted to complex road topologies, meeting the demands of vehicle navigation under diverse road conditions.

\section{Ablation Study}
We evaluated the effectiveness of the Tamba module. For comparison, we used both the traditional attention mechanism module and the Mamba module on the validation set. In addition, we assessed the efficacy of our proposed pedestrian-traffic light joint polyline encoding strategy. Table \ref{tab:methods_performance_joint} presents the performance comparison of our proposed three methods (including Attention variant, Mamba variant, and Tamba variant) under conditions with and without joint polyline encoding (w/ joint vs. w/o joint). Specifically, the table lists the performance of each method on two evaluation metrics: minFDE and minADE, considering both six predicted trajectories (minFDE\textsubscript{6} and minADE\textsubscript{6}) and one predicted trajectory (minFDE\textsubscript{1} and minADE\textsubscript{1}).

\section{Conclusion}
This paper introduces the Tamba model, an efficient motion prediction model for the real-time and computational needs of autonomous driving. By employing a state space model (SSM) and a novel attention Tamba encoder-decoder, we reduce computational complexity and parameter count while maintaining state-of-the-art accuracy. Experiments on the Argoverse 1 and 2 datasets show that Tamba achieves superior inference speed and efficiency, with a fourfold reduction in FLOPs and over 40\% fewer parameters compared to transformer-based models. These advantages make Tamba highly suitable for high-dynamic environments requiring rapid responses.

\section{Acknowledgement}
This work is supported in part by the Europe Eureka Intelligence to Drive | Move-Save-Win project (with funding from the UKRI Innovate UK project under Grant No. 10071278) as well as the Horizon Europe COVER project, No. 101086228 (with funding from UKRI grant EP/Y028031/1). Kezhi Wang would like to acknowledge the support in part by the Royal Society Industry Fellow scheme.

{
    \small
    \bibliographystyle{ieeenat_fullname}

\begin{thebibliography}{41}
\providecommand{\natexlab}[1]{#1}
\providecommand{\url}[1]{\texttt{#1}}
\expandafter\ifx\csname urlstyle\endcsname\relax
  \providecommand{\doi}[1]{doi: #1}\else
  \providecommand{\doi}{doi: \begingroup \urlstyle{rm}\Url}\fi

\bibitem[Carion et~al.(2020)Carion, Massa, Synnaeve, Usunier, Kirillov, and Zagoruyko]{carion2020end}
Nicolas Carion, Francisco Massa, Gabriel Synnaeve, Nicolas Usunier, Alexander Kirillov, and Sergey Zagoruyko.
\newblock End-to-end object detection with transformers.
\newblock In \emph{European conference on computer vision (ECCV)}, pages 213--229. Springer, 2020.

\bibitem[Chang et~al.(2019)Chang, Lambert, Sangkloy, Singh, Bak, Hartnett, Wang, Carr, Lucey, Ramanan, and Hays]{Argoverse}
Ming-Fang Chang, John~W Lambert, Patsorn Sangkloy, Jagjeet Singh, Slawomir Bak, Andrew Hartnett, De Wang, Peter Carr, Simon Lucey, Deva Ramanan, and James Hays.
\newblock Argoverse: 3d tracking and forecasting with rich maps.
\newblock In \emph{Conference on Computer Vision and Pattern Recognition (CVPR)}, 2019.

\bibitem[Cheng et~al.(2024)Cheng, Zhu, Wang, Hao, Liu, Cheng, Wang, and Chang]{cheng2024ivgaze}
Yihua Cheng, Yaning Zhu, Zongji Wang, Hongquan Hao, Yongwei Liu, Shiqing Cheng, Xi Wang, and Hyung~Jin Chang.
\newblock What do you see in vehicle? comprehensive vision solution for in-vehicle gaze estimation.
\newblock In \emph{Proceedings of the IEEE/CVF Conference on Computer Vision and Pattern Recognition (CVPR)}, 2024.

\bibitem[Cui et~al.(2023)Cui, Casas, Wong, Suo, and Urtasun]{cui2023gorela}
Alexander Cui, Sergio Casas, Kelvin Wong, Simon Suo, and Raquel Urtasun.
\newblock Gorela: Go relative for viewpoint-invariant motion forecasting.
\newblock In \emph{2023 IEEE International Conference on Robotics and Automation (ICRA)}, pages 7801--7807. IEEE, 2023.

\bibitem[Gao et~al.(2020)Gao, Sun, Zhao, Shen, Anguelov, Li, and Schmid]{gao2020vectornet}
Jiyang Gao, Chen Sun, Hang Zhao, Yi Shen, Dragomir Anguelov, Congcong Li, and Cordelia Schmid.
\newblock Vectornet: Encoding hd maps and agent dynamics from vectorized representation.
\newblock In \emph{Proceedings of the IEEE/CVF conference on computer vision and pattern recognition}, pages 11525--11533, 2020.

\bibitem[Gilles et~al.(2022)Gilles, Sabatini, Tsishkou, Stanciulescu, and Moutarde]{gilles2022thomas}
Thomas Gilles, Stefano Sabatini, Dzmitry Tsishkou, Bogdan Stanciulescu, and Fabien Moutarde.
\newblock Thomas: Trajectory heatmap output with learned multi-agent sampling.
\newblock In \emph{International Conference on Learning Representations (ICLR)}, 2022.

\bibitem[Gu and Dao(2023)]{gu2023mamba}
Albert Gu and Tri Dao.
\newblock Mamba: Linear-time sequence modeling with selective state spaces.
\newblock \emph{arXiv preprint arXiv:2312.00752}, 2023.

\bibitem[Gu et~al.(2020)Gu, Dao, Ermon, Rudra, and R{\'e}]{gu2020hippo}
Albert Gu, Tri Dao, Stefano Ermon, Atri Rudra, and Christopher R{\'e}.
\newblock Hippo: Recurrent memory with optimal polynomial projections.
\newblock \emph{Advances in neural information processing systems}, 33:\penalty0 1474--1487, 2020.

\bibitem[Gu et~al.(2022)Gu, Goel, Gupta, and R{\'e}]{gu2022parameterization}
Albert Gu, Karan Goel, Ankit Gupta, and Christopher R{\'e}.
\newblock On the parameterization and initialization of diagonal state space models.
\newblock \emph{Advances in Neural Information Processing Systems}, 35:\penalty0 35971--35983, 2022.

\bibitem[Gu et~al.(2021)Gu, Sun, and Zhao]{gu2021densetnt}
Junru Gu, Chen Sun, and Hang Zhao.
\newblock Densetnt: End-to-end trajectory prediction from dense goal sets.
\newblock In \emph{Proceedings of the IEEE/CVF International Conference on Computer Vision}, pages 15303--15312, 2021.

\bibitem[Gupta et~al.(2022)Gupta, Gu, and Berant]{gupta2022diagonal}
Ankit Gupta, Albert Gu, and Jonathan Berant.
\newblock Diagonal state spaces are as effective as structured state spaces.
\newblock \emph{Advances in Neural Information Processing Systems}, 35:\penalty0 22982--22994, 2022.

\bibitem[Huang et~al.(2024)Huang, Cheng, and Wang]{huang2024efficient}
Yizhou Huang, Yihua Cheng, and Kezhi Wang.
\newblock Efficient driving behavior narration and reasoning on edge device using large language models.
\newblock \emph{arXiv preprint arXiv:2409.20364}, 2024.

\bibitem[Kothari et~al.(2021)Kothari, Sifringer, and Alahi]{kothari2021interpretable}
Parth Kothari, Brian Sifringer, and Alexandre Alahi.
\newblock Interpretable social anchors for human trajectory forecasting in crowds.
\newblock In \emph{Proceedings of the IEEE/CVF Conference on Computer Vision and Pattern Recognition}, pages 15556--15566, 2021.

\bibitem[Lambert and Hays(2021)]{TrustButVerify}
John Lambert and James Hays.
\newblock Trust, but verify: Cross-modality fusion for hd map change detection.
\newblock In \emph{Proceedings of the Neural Information Processing Systems Track on Datasets and Benchmarks (NeurIPS Datasets and Benchmarks 2021)}, 2021.

\bibitem[Lee et~al.(2016)Lee, Purushwalkam Shiva~Prakash, Cogswell, Ranjan, Crandall, and Batra]{lee2016stochastic}
Stefan Lee, Senthil Purushwalkam Shiva~Prakash, Michael Cogswell, Viresh Ranjan, David Crandall, and Dhruv Batra.
\newblock Stochastic multiple choice learning for training diverse deep ensembles.
\newblock \emph{Advances in Neural Information Processing Systems}, 29, 2016.

\bibitem[Liang et~al.(2020)Liang, Yang, Hu, Chen, Liao, Feng, and Urtasun]{liang2020learning}
Ming Liang, Bin Yang, Rui Hu, Yun Chen, Renjie Liao, Song Feng, and Raquel Urtasun.
\newblock Learning lane graph representations for motion forecasting.
\newblock In \emph{Computer Vision--ECCV 2020: 16th European Conference, Glasgow, UK, August 23--28, 2020, Proceedings, Part II 16}, pages 541--556. Springer, 2020.

\bibitem[Liu et~al.(2024)Liu, Cheng, Chen, Broszio, Li, Zhao, Sester, and Yang]{Liu_2024_CVPR}
Mengmeng Liu, Hao Cheng, Lin Chen, Hellward Broszio, Jiangtao Li, Runjiang Zhao, Monika Sester, and Michael~Ying Yang.
\newblock Laformer: Trajectory prediction for autonomous driving with lane-aware scene constraints.
\newblock In \emph{Proceedings of the IEEE/CVF Conference on Computer Vision and Pattern Recognition (CVPR) Workshops}, pages 2039--2049, 2024.

\bibitem[Liu et~al.(2021)Liu, Yan, and Alahi]{liu2021social}
Yuejiang Liu, Qi Yan, and Alexandre Alahi.
\newblock Social nce: Contrastive learning of socially-aware motion representations.
\newblock In \emph{Proceedings of the IEEE/CVF International Conference on Computer Vision}, pages 15118--15129, 2021.

\bibitem[Mangalam et~al.(2021)Mangalam, An, Girase, and Malik]{mangalam2021goals}
Karttikeya Mangalam, Yang An, Harshayu Girase, and Jitendra Malik.
\newblock From goals, waypoints \& paths to long term human trajectory forecasting.
\newblock In \emph{Proceedings of the IEEE/CVF International Conference on Computer Vision}, pages 15233--15242, 2021.

\bibitem[Mohamed et~al.(2020)Mohamed, Qian, Elhoseiny, and Claudel]{mohamed2020social}
Abduallah Mohamed, Kun Qian, Mohamed Elhoseiny, and Christian Claudel.
\newblock Social-stgcnn: A social spatio-temporal graph convolutional neural network for human trajectory prediction.
\newblock In \emph{Proceedings of the IEEE/CVF conference on computer vision and pattern recognition}, pages 14424--14432, 2020.

\bibitem[Nayakanti et~al.(2023)Nayakanti, Al-Rfou, Zhou, Goel, Refaat, and Sapp]{nayakanti2023wayformer}
Nigamaa Nayakanti, Rami Al-Rfou, Aurick Zhou, Kratarth Goel, Khaled~S Refaat, and Benjamin Sapp.
\newblock Wayformer: Motion forecasting via simple \& efficient attention networks.
\newblock In \emph{2023 IEEE International Conference on Robotics and Automation (ICRA)}, pages 2980--2987. IEEE, 2023.

\bibitem[Ngiam et~al.(2022)Ngiam, Vasudevan, Caine, Zhang, Chiang, Ling, Roelofs, Bewley, Liu, Venugopal, et~al.]{ngiamscene}
Jiquan Ngiam, Vijay Vasudevan, Benjamin Caine, Zhengdong Zhang, Hao-Tien~Lewis Chiang, Jeffrey Ling, Rebecca Roelofs, Alex Bewley, Chenxi Liu, Ashish Venugopal, et~al.
\newblock Scene transformer: A unified architecture for predicting future trajectories of multiple agents.
\newblock In \emph{2023 IEEE International Conference on Robotics and Automation (ICRA)}, 2022.

\bibitem[Salzmann et~al.(2020)Salzmann, Ivanovic, Chakravarty, and Pavone]{salzmann2020trajectron++}
Tim Salzmann, Boris Ivanovic, Punarjay Chakravarty, and Marco Pavone.
\newblock Trajectron++: Dynamically-feasible trajectory forecasting with heterogeneous data.
\newblock In \emph{Computer Vision--ECCV 2020: 16th European Conference, Glasgow, UK, August 23--28, 2020, Proceedings, Part XVIII 16}, pages 683--700. Springer, 2020.

\bibitem[Sherstinsky(2020)]{sherstinsky2020fundamentals}
Alex Sherstinsky.
\newblock Fundamentals of recurrent neural network (rnn) and long short-term memory (lstm) network.
\newblock \emph{Physica D: Nonlinear Phenomena}, 404:\penalty0 132306, 2020.

\bibitem[Shi et~al.(2023)Shi, Wang, Long, Zhou, Tang, Zheng, and Hua]{shi2023representing}
Liushuai Shi, Le Wang, Chengjiang Long, Sanping Zhou, Wei Tang, Nanning Zheng, and Gang Hua.
\newblock Representing multimodal behaviors with mean location for pedestrian trajectory prediction.
\newblock \emph{IEEE transactions on pattern analysis and machine intelligence}, 2023.

\bibitem[Shi et~al.(2024)Shi, Jiang, Dai, and Schiele]{shi2024mtr++}
Shaoshuai Shi, Li Jiang, Dengxin Dai, and Bernt Schiele.
\newblock Mtr++: Multi-agent motion prediction with symmetric scene modeling and guided intention querying.
\newblock \emph{IEEE Transactions on Pattern Analysis and Machine Intelligence}, 2024.

\bibitem[Smith et~al.(2024)Smith, De~Mello, Kautz, Linderman, and Byeon]{smith2024convolutional}
Jimmy Smith, Shalini De~Mello, Jan Kautz, Scott Linderman, and Wonmin Byeon.
\newblock Convolutional state space models for long-range spatiotemporal modeling.
\newblock \emph{Advances in Neural Information Processing Systems}, 36, 2024.

\bibitem[Su et~al.(2022)Su, Wang, and Yang]{su2022qml}
Tong Su, Xishun Wang, and Xiaodong Yang.
\newblock Qml for argoverse 2 motion forecasting challenge.
\newblock \emph{arXiv preprint arXiv:2207.06553}, 2022.

\bibitem[Varadarajan et~al.(2022)Varadarajan, Hefny, Srivastava, Refaat, Nayakanti, Cornman, Chen, Douillard, Lam, Anguelov, et~al.]{varadarajan2022multipath++}
Balakrishnan Varadarajan, Ahmed Hefny, Avikalp Srivastava, Khaled~S Refaat, Nigamaa Nayakanti, Andre Cornman, Kan Chen, Bertrand Douillard, Chi~Pang Lam, Dragomir Anguelov, et~al.
\newblock Multipath++: Efficient information fusion and trajectory aggregation for behavior prediction.
\newblock In \emph{2022 International Conference on Robotics and Automation (ICRA)}, pages 7814--7821. IEEE, 2022.

\bibitem[Vaswani et~al.(2017)Vaswani, Shazeer, Parmar, Uszkoreit, Jones, Gomez, Kaiser, and Polosukhin]{vaswani2017attention}
Ashish Vaswani, Noam Shazeer, Niki Parmar, Jakob Uszkoreit, Llion Jones, Aidan~N Gomez, {\L}ukasz Kaiser, and Illia Polosukhin.
\newblock Attention is all you need.
\newblock \emph{Advances in neural information processing systems}, 30, 2017.

\bibitem[Wang et~al.(2023{\natexlab{a}})Wang, Zhu, Yu, Li, Ma, Jin, Ren, Ren, Wang, and Yang]{wang2023ganet}
Mingkun Wang, Xinge Zhu, Changqian Yu, Wei Li, Yuexin Ma, Ruochun Jin, Xiaoguang Ren, Dongchun Ren, Mingxu Wang, and Wenjing Yang.
\newblock Ganet: Goal area network for motion forecasting.
\newblock In \emph{2023 IEEE International Conference on Robotics and Automation (ICRA)}, pages 1609--1615. IEEE, 2023{\natexlab{a}}.

\bibitem[Wang et~al.(2023{\natexlab{b}})Wang, Chen, Lertniphonphan, Chen, Bao, Zheng, Zhang, Huang, and Zhang]{wang2023technical}
Zhepeng Wang, Feng Chen, Kanokphan Lertniphonphan, Siwei Chen, Jinyao Bao, Pengfei Zheng, Jinbao Zhang, Kaer Huang, and Tao Zhang.
\newblock Technical report for argoverse challenges on unified sensor-based detection, tracking, and forecasting.
\newblock \emph{arXiv preprint arXiv:2311.15615}, 2023{\natexlab{b}}.

\bibitem[Wilson et~al.(2021)Wilson, Qi, Agarwal, Lambert, Singh, Khandelwal, Pan, Kumar, Hartnett, Pontes, Ramanan, Carr, and Hays]{Argoverse2}
Benjamin Wilson, William Qi, Tanmay Agarwal, John Lambert, Jagjeet Singh, Siddhesh Khandelwal, Bowen Pan, Ratnesh Kumar, Andrew Hartnett, Jhony~Kaesemodel Pontes, Deva Ramanan, Peter Carr, and James Hays.
\newblock Argoverse 2: Next generation datasets for self-driving perception and forecasting.
\newblock In \emph{Proceedings of the Neural Information Processing Systems Track on Datasets and Benchmarks (NeurIPS Datasets and Benchmarks 2021)}, 2021.

\bibitem[Wong et~al.(2024)Wong, Xia, Zou, Wang, and You]{Wong_2024_CVPR}
Conghao Wong, Beihao Xia, Ziqian Zou, Yulong Wang, and Xinge You.
\newblock Socialcircle: Learning the angle-based social interaction representation for pedestrian trajectory prediction.
\newblock In \emph{Proceedings of the IEEE/CVF Conference on Computer Vision and Pattern Recognition (CVPR)}, pages 19005--19015, 2024.

\bibitem[Ye et~al.(2021)Ye, Cao, and Chen]{ye2021tpcn}
Maosheng Ye, Tongyi Cao, and Qifeng Chen.
\newblock Tpcn: Temporal point cloud networks for motion forecasting.
\newblock In \emph{Proceedings of the IEEE/CVF Conference on Computer Vision and Pattern Recognition}, pages 11318--11327, 2021.

\bibitem[Yuan et~al.(2021)Yuan, Weng, Ou, and Kitani]{yuan2021agentformer}
Ye Yuan, Xinshuo Weng, Yanglan Ou, and Kris~M Kitani.
\newblock Agentformer: Agent-aware transformers for socio-temporal multi-agent forecasting.
\newblock In \emph{Proceedings of the IEEE/CVF International Conference on Computer Vision}, pages 9813--9823, 2021.

\bibitem[Zhang et~al.(2025)Zhang, Liu, Reid, Hartley, Zhuang, and Tang]{zhang2025motion}
Zeyu Zhang, Akide Liu, Ian Reid, Richard Hartley, Bohan Zhuang, and Hao Tang.
\newblock Motion mamba: Efficient and long sequence motion generation.
\newblock In \emph{European Conference on Computer Vision}, pages 265--282. Springer, 2025.

\bibitem[Zhou et~al.(2024)Zhou, Lin, Shan, Wang, Sun, and Yang]{zhou2024drivinggaussian}
Xiaoyu Zhou, Zhiwei Lin, Xiaojun Shan, Yongtao Wang, Deqing Sun, and Ming-Hsuan Yang.
\newblock Drivinggaussian: Composite gaussian splatting for surrounding dynamic autonomous driving scenes.
\newblock In \emph{Proceedings of the IEEE/CVF Conference on Computer Vision and Pattern Recognition}, pages 21634--21643, 2024.

\bibitem[Zhou et~al.(2022)Zhou, Ye, Wang, Wu, and Lu]{zhou2022hivt}
Zikang Zhou, Luyao Ye, Jianping Wang, Kui Wu, and Kejie Lu.
\newblock Hivt: Hierarchical vector transformer for multi-agent motion prediction.
\newblock In \emph{Proceedings of the IEEE/CVF Conference on Computer Vision and Pattern Recognition}, pages 8823--8833, 2022.

\bibitem[Zhou et~al.(2023)Zhou, Wang, Li, and Huang]{zhou2023query}
Zikang Zhou, Jianping Wang, Yung-Hui Li, and Yu-Kai Huang.
\newblock Query-centric trajectory prediction.
\newblock In \emph{Proceedings of the IEEE/CVF Conference on Computer Vision and Pattern Recognition}, pages 17863--17873, 2023.

\bibitem[Zhu et~al.(2024)Zhu, Liao, Zhang, Wang, Liu, and Wang]{zhu2024vision}
Lianghui Zhu, Bencheng Liao, Qian Zhang, Xinlong Wang, Wenyu Liu, and Xinggang Wang.
\newblock Vision mamba: Efficient visual representation learning with bidirectional state space model.
\newblock \emph{arXiv preprint arXiv:2401.09417}, 2024.

\end{thebibliography}

}

\end{document}